\pdfoutput=1

\documentclass[11pt]{article}

\usepackage[]{ACL2023}

\usepackage{times}
\usepackage{latexsym}
\usepackage[colorinlistoftodos,textsize=tiny,disable]{todonotes}
\usepackage{xcolor}
\usepackage[T1]{fontenc}

\usepackage[utf8]{inputenc}

\usepackage{microtype}

\usepackage{inconsolata}

\usepackage{enumitem}

\newcommand{\stitle}[1]{\textbf{#1}}

\title{On Context-aware Detection of Cherry-picking in News Reporting}

\author{Israa Jaradat , Haiqi Zhang, Chengkai Li \\
         Department of Computer Science and Engineering \\
         University of Texas at Arlington \\
         Arlington, TX, USA \\ \{israa.jaradat,haiqi.zhang\}@mavs.uta.edu, cli@uta.edu}

\begin{document}
\maketitle
\begin{abstract}
Cherry-picking refers to the deliberate selection of evidence or facts that favor a particular viewpoint while ignoring or distorting evidence that supports an opposing perspective. Manually identifying cherry-picked statements in news stories can be challenging. In this study, we introduce a novel approach to detecting cherry-picked statements by identifying missing important statements in a target news story using language models and contextual information from other news sources. Furthermore, this research introduces a novel dataset specifically designed for training and evaluating cherry-picking detection models. Our best performing model achieves an F-1 score of about 89\% in detecting important statements. Moreover, results show the effectiveness of incorporating external knowledge from alternative narratives when assessing statement importance.
\end{abstract}

\section{Introduction}

\label{sec:introduction}
The erosion of citizens' trust in the media has reached a critical stage, posing significant harm to the foundations of democracy~\cite{dan2021visual, druckman2005impact, chiang2011media}. Research shows evidence of a decline in media trust among citizens ~\cite{stromback2020news}. 
Media losing trust can be attributed to the perception of political bias in media reporting~\cite{lee2010they} and consumers' exposure to disinformation~\cite{lee2023antecedents, hameleers2022whom}.

\emph{Cherry-picking}, a widely used but often imperceptible practice in news reporting, is a device that aids bias and disinformation~\cite{vincent2019global, paul2019thinker}. 
It is defined as a logical fallacy that involves deliberately selecting (or exaggerating) facts that bolster a certain argument and omitting (or under-emphasizing) facts that support the opposing viewpoint~\cite{joelbest}. Its common aliases in literature include \textit{one-sideness}~\cite{paul2006detect}, \textit{argument by half-truth}~\cite{sproule2001authorship}, and \textit{case-making}~\cite{lee1945analysis}.

In news reporting, cherry-picking can be observed across scales as large as in selecting which events to cover, and as subtle as in choosing specific words for a particular story. Furthermore, cherry-picking can manifest in various types of data such as charts ~\cite{asudeh2020detecting} or when particular statistics~\cite{joelbest} are chosen to bolster a certain claim~\cite{wu2017computational}.

This study's focus is cherry-picking statements within news articles, particularly when only certain statements are included while others are omitted. A multitude of media outlets, including prominent and highly regarded ones, exhibit biases in various topics within domains such as politics and science~\cite{baron2006persistent}. These media outlets often exercise caution in curating the statements included in their stories covering controversial topics~\cite{paul2019thinker, paul2006detect}. This selection of statements is capable of influencing public opinion on pivotal subjects, such as climate change, vaccination, or elections, by misleading readers to perceive partial truth as holistic truth~\cite{fleming1995understanding}. 

We introduce a novel method for identifying cherry-picking in news stories by leveraging context derived from other narratives. Our method employs language models, including fine-tuned embedding models and few/zero shot prompting of generative models, to assess statements' importance based on a given context to identify cherry-picked ones. The fine-tuned models are trained on the novel \textit{Cherry} dataset we curated. The proposed method shows promising results in inferring a statement's importance given external contextual information. Our top-performing model achieves an F-1 score of 89\% and an accuracy of 90\%.

To the best of our knowledge, this is the first study that addresses the detection of cherry-picking in text using computational approach. The key contributions of this work are as follows: (a) The formulation and modeling of the problem from a computational perspective. (b) The novel end-to-end cherry-picking detection approach that infuses contextual information with statements through different language modeling techniques.
(c) A new cherry-picking annotated dataset which can prove useful in not only detection but also other lines of research related to cherry-picking. (d) Thorough experimentation to determine the most effective context to consider when assessing cherry-picking. All artifacts of this work are released under the GNU General Public License v3.0 at our anonymous Github repository \url{https://github.com/cherry-pic/Cherry}. 

\section{Related Work} 
\label{sec:related_work}
Research on media bias can be classified into three levels based on text granularity: word-level, sentence-level, and article-level~\cite{chen-etal-2020-analyzing}. Distinguishing between these levels is not always straightforward, as they often intersect and influence each other. For instance, word-level bias is dependent on sentence or article context, and similarly sentence-level bias is contingent on the broader article context.

\noindent \textbullet\hspace{1mm} {\emph{Word-level bias.}} 
The choice of words in news articles can influence the audience's opinions towards a particular event~\cite{hamborg-2020-media}. Word-level bias encompasses two primary categories: framing bias and epistemological bias~\cite{recasens2013linguistic}. The connotations associated with the same word can vary by the specific context in which they are employed~\cite{spinde2021automated}.
 
\noindent \textbullet\hspace{1mm} {\emph{Sentence-level bias.}} Numerous studies focus on detecting bias at sentence level, considering it as the fundamental level that can be combined with article-level bias~\cite{spinde-etal-2021-neural-media}. 
Using lexicon of bias words, \citet{hube2018detecting} introduced a method for identifying biased statements from Wikipedia. In \citet{lei-etal-2022-sentence}, biased sentences were identified in news articles, serving to explain the overall bias present throughout articles. Focusing on media's sentiment toward target entities, \citet{fan-etal-2019-plain} examined sentence-level bias along with article context. 

\noindent \textbullet\hspace{1mm} {\emph{Article-level bias.}} 
\citet{baly-etal-2020-detect} constructed a dataset of news articles annotated with left, center, and right leanings, then used machine learning models to predict articles' political ideology.

Prior research on media bias primarily focused on predisposition towards particular viewpoints, 
leaving a notable gap in exploring media bias related to cherry-picking in news articles. 
Some studies have examined the impact of cherry-picking in computational fact-checking~\cite{wu2017computational, lin2021detecting} and the detection of cherry-picking in trendlines~\cite{asudeh2020detecting}. Nevertheless, there has been no specific investigation into end-to-end cherry-picking detection in news articles. 




\section{Cherry-picking Facts in News Stories: Problem Definition and Modeling}
\label{sec:cherry-picking_definition}
Media outlets cherry-pick in news reporting by prioritizing less crucial statements that align with their biases to appear in their narrative, or neglecting more important relevant statements, especially those against their biases \citep{paul2006detect, vincent2019global, brown1963techniques}. 
To offer a more illustrative portrayal of cherry-picked statements, Table~\ref{table:juxtaposition} in~\ref{sec:juxta_example} provides a juxtaposition of three authentic news stories from distinctively biased outlets covering the same event. 

The example in Table~\ref{table:juxtaposition}, specifically rows 3, 7, and 8 demonstrates two important insights that guide our approach. \todo[color=green]{Can we point to specific places in the table to explain what we are saying here? If speace allows, you can move back the discussion from Appendix to here.} \emph{First}, cherry-picked statements can be potentially discerned by their importance to the event being covered. Therefore, we exploit statement importance in detecting cherry-picking. We opt for assessing cherry-picking on the statement (i.e., sentence) level since it is a common practice in news reporting to represent a singular idea in one sentence \cite{mencher1997news}. Note that stories often face constraints on space due to factors such as reader's attention span, which results in prioritizing the most important statements to appear in an article. 
\emph{Second}, the importance of a statement is relevant, contextualized by whether it is discussed or not in other stories about the same event---simply put, together all stories depict a comprehensive picture. For a comprehensive and representative context, it is crucial to incorporate narratives from other sources with different biases~\cite{institute1939fine}. 

We categorize cherry-picking into two forms: \emph{overemphasis} and \emph{underemphasis}. \todo[color=yellow]{Several problems with regard to under- vs. over-emphasis. These are worth pondering after paper submission. 1) Our dataset is believed to have very few examples of overempahsis. It is unclear whether it is due to the particular events or particular ways of collecting data. Or whether it is a generally applicable observation. 2) and we didn't explain what reasons made underemphasis more prevalent, regardless of whether it is due to the data specifically or generally applicable.  3) if it is general, are there studies that agree with our claim?} Overemphasis occurs when the coverage of a particular event focuses on non-essential statements and magnifies them, creating the impression that they are significant components of the event~\cite{fleming1995understanding, lee1945analysis}. Conversely, underemphasis occurs when a story entirely excludes or slants a crucial fact from an event coverage \citep{sproule2001authorship, lakomy2020between}. 

The detection of overemphasis requires not only evaluating a statement's importance and presence but also analyzing other factors, such as exaggeration cues, sentiment, opinion, in addition to measuring repetition and the amount of space allocated to the statements within a report. Studying the presence of a statement requires a different approach than studying its absence. For instance, detecting exaggeration requires computing lexical features which is not applicable to the analysis of textual absence since there is no text to begin with. We will address cherry-picking by overemphasis in future work, and we focus on cherry-picking by underemphasis in this study. \todo[color=pink]{The analysis of overemphasis vs underemphasis is unconvincing. Something present in one article can be missing in another; vice versa. Given their opposite nature, your explanantion doesn't make a lot of sense to me.} 
To identify underemphasis, analysis of textual absence using context from other narratives and sources is required.  
To find important statements in an event, one needs to read stories from other distinctively biased sources, then evaluate the importance of each statement within the story based on their understanding of the event.

Drawing from the discussion above, the problem of detecting cherry-picking through under-emphasis is formulated as follows.
Consider an event $e$$=$$\{d_1, ..., d_n\}$ comprising various narratives (i.e., documents). Each document $d_i$$=$$\{s_1, ... , s_m\}$ contains a collection of statements (i.e., sentences, as we do not consider multiple sentences collectively as a single statement in this study). The statements from all documents in $e$ collectively form the universal set of statements $S_e$$=$$\{$$d_1 \cup  d_2 ... \cup  d_n$\} for the entire event. Our goal is to determine the set of important statements that are missing from each document, i.e., $c_i$$=$$I_e-d_i$ in which $I_e \subset  S_e$ represents the important statements regarding $e$, among all statements. The problem then reduces to finding $I_e$ based on the event's context. In this study we use documents from distinctively biased sources covering event $e$ as its context. This is based on the need for alternative narratives (i.e., the other perspective) to assess cherry-picking ~\cite{institute1939fine}. 
To determine outlets biases, we utilized categorization of media outlet bias based on unanimous agreement from three sources---Media Bias Fact Check (MBFC),\footnote{\url{https://mediabiasfactcheck.com/}} AllSides,\footnote{\url{https://www.allsides.com/}} and Ad Fontes Media.\footnote{\url{https://adfontesmedia.com/}} These sources are widely utilized in media bias analysis, providing researchers with tools and ratings to evaluate the political bias and credibility of news sources. The details of how they assess bias in news outlets can be found in~\ref{sec:rating_method}.
Finally, we utilize $c_i$ to assess cherry-picking of a given outlet by counting the number of cherry-picked statements in each document $d_i$ from the outlet, followed by averaging this number across a time span of events.

\section{Dataset}
\label{sec:dataset}

To train and evaluate our models for gauging statement importance given event context, we curated a novel cherry-picking detection dataset \textit{Cherry}. 
Our dataset is composed of 3,346 examples in total. Every example contains a statement $s \in S_e$ (i.e., a single sentence from a news report of event $e$), an event context $d \in e$ which contains an article covering the event collected from another source, and an importance label $Y$ that indicates whether the statement is important to the event or not.

\subsection{Data Collection} 
To generate the examples in the dataset, we compiled a list of 41 noteworthy news sources, such as CNN.com, Reuters.com, and FoxNews.com, selected for their high publication frequency and distinct political biases as determined by the ratings from the aforementioned three websites.\footnote{The full list of the 41 sources can be found at \url{https://github.com/cherry-pic/Cherry}.} Subsequently, we employed GDELT's API ~\cite{leetaru2013gdelt} to gather all news articles except for opinion articles and editorials from the chosen sources. The collected news articles cover the time span between December 2019 and January 2021. 

Next, we used DBSCAN~\cite{ester1996density} to cluster the collected articles into events based on the cosine similarity calculated between the vectors of the articles, with each article being vectorized through the concatenation of BERT~\cite{devlin-etal-2019-bert} and TF-IDF representations derived from its headline and initial paragraph. We set DBSCAN's parameters to 0.04 for the radius of neighborhood around data points ($\varepsilon$) and 2 for the minimum cluster size.

After a manual inspection of the generated clusters, we curated a subset of 82 clusters, each corresponding to a distinct controversial event. Examples of such events include the January 6th, 2021 U.S. Capitol attack and the alleged foreign intervention in the 2020 U.S. presidential elections. ~\footnote{The full list of the 82 events is available at \url{https://github.com/cherry-pic/Cherry}.}

For each event, we used NLTK~\cite{bird2009natural} sentence tokenizer to segment the articles into statements, and then clustered the statements based on semantic similarity. Similar to how articles were clustered, we used BERT and TF-IDF to represent statements and applied DBSCAN with an value of 0.07 for $\varepsilon$, a minimum cluster size of 2, and cosine similarity of statement vectors as the similarity function. Table~\ref{table:clustering} in~\ref{sec:clust_example} shows a sample of clustered statements within a single event. Statement clustering facilitates data augmentation on the dataset, where the collected label for a single statement was cast over all statements in the same cluster. As a result, multiple examples were labeled with a single example labeling effort. 

\begin{figure}[hbt]
\centering
\includegraphics[width=0.45\textwidth]{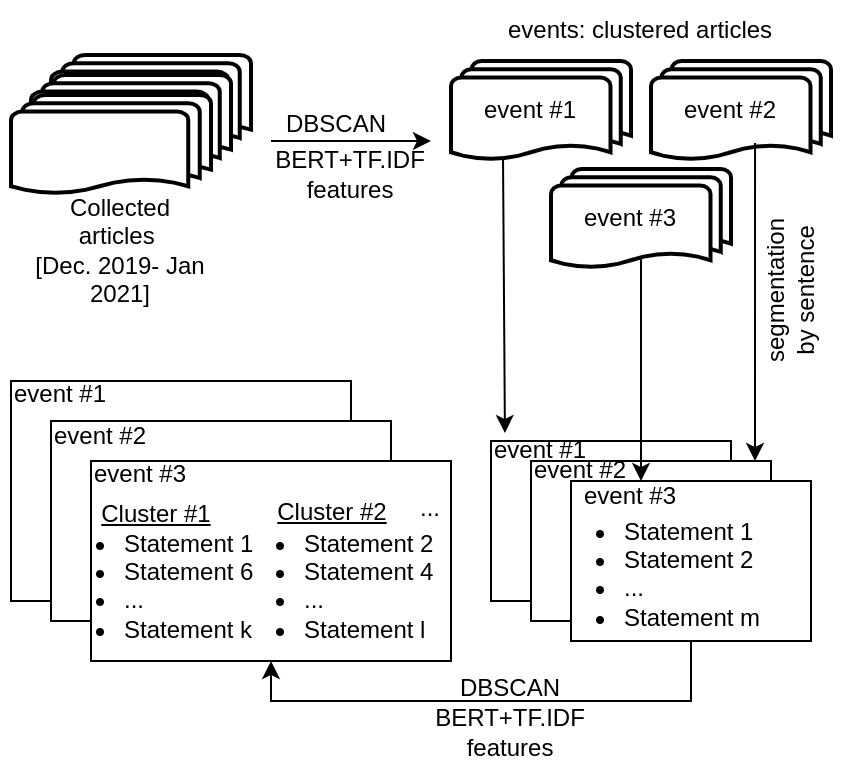}
\caption{Data collection and preparation pipeline.}
\vspace{-2mm}
\label{data_collection}
\end{figure}

The data collection, clustering, and segmentation pipeline is depicted in Figure \ref{data_collection}. After clustering the statements, we fed them along with their respective context into our custom data annotation tool, as explained in Section~\ref{subsec:data_annotation}. 

\subsection{Data Annotation}
\label{subsec:data_annotation} 
We developed a data annotation tool~\footnote{Public URL withheld for anonymity.} to collect labels for statements that reflect the statements' importance to corresponding events. 
The annotators comprise one faculty member and nine Ph.D. students from a university research lab. They are diverse in terms of cultural background, nationality, gender, and age.\todo[color=yellow]{The data annotation website is not accessible. At least for arxiv, we need to check its status regularly to make sure it is available.} 
The ensuing discussion in this section provides a brief explanation of the data annotation tool and process, while more details can be found in~\ref{sec:interface}. 

For each event, the annotators are presented with a news article from a least-biased source (e.g., Reuters.com), along with clusters of statements related to the event, as shown in Figure~\ref{cherry_web} in~\ref{sec:interface}.  The annotators are required to thoroughly read the articles and gain a comprehensive understanding of the event covered. They are instructed to assign one of the following five labels to the cluster: (1) \textit {very important}, (2) \textit {kind of important}, (3) \textit{not very important}, (4) \textit{the excerpts might be incorrect}, and (5) \textit{I am not sure}. These labels convey the perceived level of importance of the statements in the cluster. Following the collection of labels from annotators, we filtered out examples with less than 3 annotators or less than 75\% agreement ratio (i.e., number of majority votes divided by the total number of annotators for an example).  

\subsection{Dataset Statistics}
Considering the inherent difficulty in delineating the decision boundary between the different labels described in Section~\ref{subsec:data_annotation}), we opted to explore various binary and multinomial classification configurations as depicted in Table \ref{table:class_distrib}. This involved merging certain labels into a single class or excluding specific labels from the dataset. The class distribution for each of the four classification configurations is provided in Table \ref{table:class_distrib}. For each configuration, we split the dataset randomly into 85\% for training and 15\% for testing.

Each example in the dataset consists of labels, context collected from other articles, and statements. Multiple statements stem from the same event, and consequently the context repeats within the dataset examples. This leads to half of the examples in the dataset sharing the same context. Hence, we split the dataset between testing and training by events instead of stratification to avoid any data leakage (i.e., model's exposure to event contexts from the testing dataset during training).

\begin{table}[hbt!]
\small
\centering
\begin{tabular}{|p{0.25in}| p{0.6in} p{0.9in} p{0.6in}|}
\hline
\textbf{Conf.} & \textbf{Class 1} & \textbf{Class 2} & \textbf{Class 3}\\ 
\hline
1 & \{(1)\} & \{(2),(3),(4),(5)\} & -           \\ 
  & 2175 (64\%) & 1232 (36\%) & - \\
\hline
2 & \{(1)\} & \{(2),(3)\}        & -           \\
  & 2175 (65\%) & 1171 (35\%)  & - \\
\hline
3 & \{(1)\}& \{(2),(3)\} & \{(4),(5)\} \\
  & 2175 (64\%) & 1171 (34\%) & 61 (2\%) \\
\hline
4 & \{(1)\} & \{(2)\} & \{(3)\}    \\
  & 2175 (65\%) & 667 (20\%) & 504 (15\%)\\
\hline
\end{tabular}
\caption{Label combinations and class distribution (number of examples and ratio with regard to the whole dataset) for each of the four classification configurations. The label for each index from (1) to (5) can be found in Section~\ref{subsec:data_annotation}.}
\label{table:class_distrib}
\end{table}
\vspace{-1mm}



\section{Models}
\label{sec:model}
We develop cherry-picking models that, given an event and a statement, estimate an importance score for the statement with regard to the event. Our models emulate the approach taken by human annotators---reading the context about the event and then assessing the statement's importance considering the context. This section describes how various techniques can be employed for this task,  including supervised, unsupervised, and zero-shot demonstration methods. 

\subsection{Fine-tuned Supervised Models}
To consider context while estimating a statement's importance, we fine-tuned supervised language models using the sequence pair classification task on our dataset (refer to Figure~\ref{fig:archi} in~\ref{sec:archi} for a detailed illustration). A sequence pair consists of a  statement and the corresponding context. The context is taken from other articles within the event collection. If the token limit of a certain language model is exceeded, the sequence is right-truncated. The encoded sequence is represented as a vector obtained from the output of the [CLS] token. This vector is subsequently passed through a linear layer to produce a class probability.

We employed both BERT~\cite{devlin-etal-2019-bert} and Longformer~\cite{beltagy2020longformer} as our supervised models.
Unlike BERT, Longformer exhibits linear scaling with input sequence length. As a result, it enables the processing of lengthier documents, accommodating approximately 4096 tokens compared to BERT's limit of 512 tokens. This capability aligns well with our requirement for typically long input sequences of context (i.e., full news articles). The primary factor that distinguishes Longformer as a computationally efficient variant from BERT is its novel self-attention mechanism. Longformer applies self-attention over a sliding window or dilated window of tokens, instead of every token in the input sequence. Additionally, to learn task-specific representations, Longformer uses global attention on preselected tokens form the input sequence. These tokens attend to every other token in the sequence, and every token in the sequence attends only to these preselected tokens, which facilitates the learning of task-specific representations~\cite{beltagy2020longformer}, further enhancing its suitability for our needs. 

\subsection{Large Language Models with Zero-shot or Few-shot Demonstration}
\label{subsec:LLMs}
The advantage of zero-shot and few-shot demonstration is that it does not require large amount of data or human supervision. In utilizing the generative language model GPT~\cite{brown2020language} to assess a statement's importance, our approach imitates the way human coders annotated the Cherry dataset. We use the same prompt in the annotation interface (i.e., the annotation instruction) with slight modifications (refer to~\ref{sec:prompting} for the full template). We first let the model read the context collected from other articles about an event. We then prompt the model to answer whether the statement is important to mention in a news story covering the event. Alternatively, a few demonstrations can be embedded within the prompt for the model to learn from, where each demonstration example contains the context article, the statement to assess, the question asking the model if the statement is important considering the context article, and the answer to the question (i.e., Yes/No). The full prompt template is included in ~\ref{sec:prompting}.

\subsection{Unsupervised Methods} 
Existing unsupervised text summarization algorithms such as LexRank~\cite{erkan2004lexrank} can facilitate finding $I_e$ (important statements regarding event $e$) with less cost and effort compared to supervised approaches.
LexRank is fit by first constructing a graph over a given text, where each sentence in the text is represented as a node. Edges between nodes are weighted based on the similarity between sentences, using common measures such as cosine similarity of TF-IDF vectors. LexRank then constructs a stochastic matrix of these weights and uses a power method to find the eigenvector centrality of each sentence. After fitting LexRank, it is given a new set of sentences from which LexRank extracts a summary by ranking the sentences based on their centrality scores which reflect their importance. Only the top $n$ (i.e., summary size) sentences appear in the summary.
For the task of cherry-picking detection, we employ LexRank to find statements' importance after fitting it (i.e., constructing its graph) on the context, then pass it a list of sentences (i.e., a news article) to rank them based on importance (i.e., extract a summary). Only sentences selected by the algorithm to appear in the summary are considered important. \todo[color=green]{If it is fit only using the context, it won't contain sentences from other places. Then how can it assign a score to any sentence outside the context? The paragraph only describes how sentences in the graph (i.e., in the context article) are scored. Not any other sentence. I don't understand how it can be fit on the context only and then summarize another article, as you described.}  \todo[color=green]{also, your description earlier said the graph is constructed using "collection of documents", not just the context}

\subsection{End-to-end Cherry-picking  Detection}
The cherry-picking detection pipeline identifies cherry-picked statements $c_i$ within a document $d_i$ given an event $e$, as follows: 
\vspace{-1em}
\begin{equation}\label{eq1}
 c_i= \{\exists s \in S_e: f(s,d) = 1\} - d_i 
\end{equation}
where $f$ is any of our aforementioned methods which assign a class of 1 for important statements and 0 for unimportant ones based on the probability score for that class.\todo{Earlier we said these methods return scores, not classes. classification threshold hasn't been mentioned till this point.} The pipeline finds $c_i$ by first segmenting all documents in event $e$ to obtain the universal set of statements $S_e$. Then each statement in $S_e$ is scored for importance using a model and the event context article $d$. Finally, each important statement is verified for its presence in each article using semantic similarity. If the statement is absent in an article, it is appended to the list of cherry-picked (i.e., underemphasized) statements associated with that article. 

\section{Experimentation and Results \footnote{All the code base, experiments and results are available at \url{https://github.com/cherry-pic/Cherry}.}}
\label{sec:experiments}

\todo{Not sure it is a good idea to use footnote for the codebase URL.}

We used the dataset described in Section \ref{sec:dataset} to evaluate the performance of the models proposed in Section \ref{sec:model}.

{\flushleft{\textbf{Context contribution to performance}}}\hspace{2mm} 
To understand the contribution of context in inferring statement importance, we utilized a BERT-base \footnote{\url{https://huggingface.co/google-bert/bert-base-uncased}} model as the baseline, with a statement alone (i.e., no context) as its input sequence.  We also created a variant of the model by forming the input sequence as the statement and its context concatenated and separated by the [SEP] token. Context in this experiment is collected from a neutral source (i.e., Reuters.com). We use macro F-1 and accuracy as models' performance measures in this and subsequent experiments. Our hypothesis is that, just as humans can make more accurate judgments regarding the importance and cherry-picking of a statement when provided with an unbiased context~\cite{institute1939fine}, models are anticipated to exhibit similar capacity. The baseline model scored 0.619 and 0.617 in accuracy and F-1, respectively. When the same model was given 100 words of context to attend to when classifying statements, its accuracy and F-1 were significantly improved to 0.846 and 0.870, respectively. 
Note that language models can still capture some signals pertinent to statements' importance from their content without context (thus better than a random guesser).

{\flushleft{\textbf{Model comparison}}}\hspace{2mm} 
Given the importance of context as established above, we compared the performance of various models at different context lengths. Toward this, we trimmed the context in both the training and test sets at different lengths measured in words, and we then fine-tuned the supervised models and evaluated all models. In addition to comparing the various proposed methods, this experiment aims to gain insights on how much context should be fed to the models. Similar to the experiment above, context in this experiment is collected from a neutral source.

In this experiment, for supervised models, we used the smaller variant BERT-base consisting of 12 transformer layers, 12 attention heads, and 110M parameters, in addition to Longformer-base \footnote{\url{https://huggingface.co/allenai/longformer-base-4096}} consisting of 12 transformer layers, 12 attention heads and 149M parameters.  We fixed the learning rate at 2e-05, the batch size at 8, and the classification threshold at 0.5. To maintain performance and efficiency, we set global attention in Longformer models at the classification and statement tokens only (refer to~\ref{sec:glob_attention} for details of experimentation on global attention locations). Additionally, we set the number of epochs for training the supervised models to 5, which required at most five minutes of training with the aforementioned parameters on an NVIDIA H100 PCIe GPU with 80 GB memory. For zero and few-shot demonstration models, we experimented with GPT, specifically \textit{gpt-3.5-turbo-16k} via OpenAI's Chat Completion API ~\footnote{\url{https://platform.openai.com/docs/guides/text-generation/chat-completions-api}} with temperature fixed at 0 for a deterministic behavior. We prompted GPT using the prompts discussed in Section~\ref{subsec:LLMs}. For unsupervised models, we used LexRank with summary size (in terms of sentences) for each event to be equal to the number of positive examples in the event from the test set, and we fixed the cosine similarity threshold at 0.1.

\begin{figure}[!htb]
\centering
\includegraphics[width=0.485\textwidth]{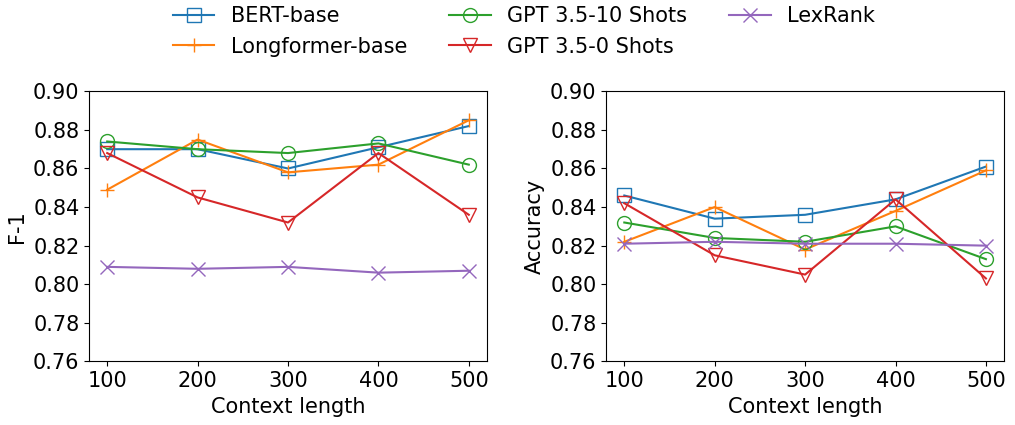}
\caption{Effect of context size (measured in words) on models' performance.}
\label{fig:model_comparison}
\end{figure}

Results in Figure~\ref{fig:model_comparison} show variation in performance as the context size increases from 100 to 500 words. The majority of the models demonstrated heightened F-1 and accuracy within a context length of 400 words.\todo[color=green]{Is it just F-1? Performance measured by ``Accuracy'' seems similar.} This amount corresponds to approximately 12 paragraphs sourced from Reuters.com. This observation indicates that a neutral news report of this specified length pertaining to a given event is sufficient for discerning instances of cherry-picking within a biased narrative. Moreover, results show that the best performing models for this task are the fine-tuned supervised models at context length of 500 words. However, only 10 demonstrations fed to the generative LLM could give decent results.

LexRank underperformed compared to other models, primarily due to its dependence on text repetition and the frequency of similar content within the document set.
Unsupervised summarizers, such as LexRank, encounter challenges in capturing important statements when they occur infrequently in the document collection.  A statement's importance is determined by its critical role as a key piece of evidence, rather than its frequency in the news articles.

For further improvement, we tested the large variant of Longformer (i.e., Longformer-large), consisting of 24 transformer layers, 16 attention heads, and 435M parameters. Longformer-large achieved the best performance compared to all other models given context size of 500, with an accuracy of 0.897 and an F-1 of 0.887.\todo[color=green]{From the figure Longformer-large is about the same as BERT-base, or even slightly worse in terms of Accuracy. The one in the figure is longformer base . This exp is talking about Longformer large which is not present in the figure} We used this model to perform the ensuing experiments. 

\begin{figure}[!htb]
\centering
\includegraphics[width=0.485\textwidth]{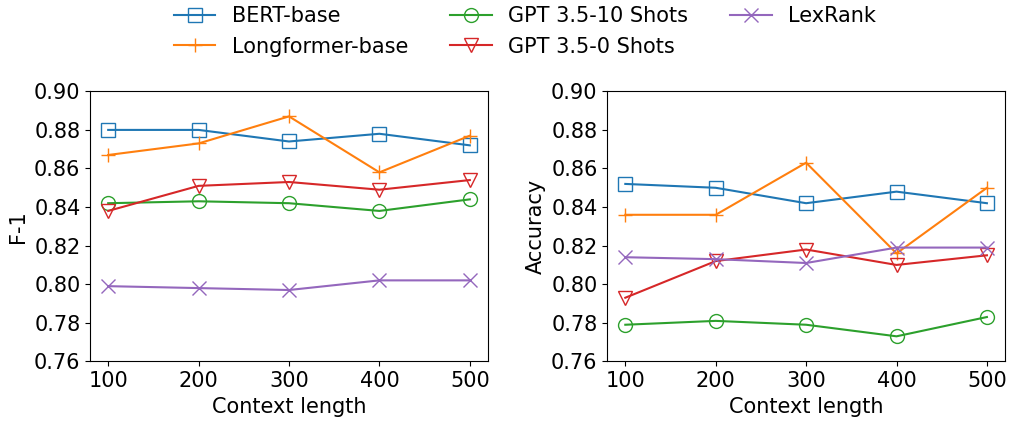}
\caption{Model performance when using a context collected and summarized in different lengths from biased news sources instead of a neutral source.}
\label{fig:biased_context}
\end{figure}

\begin{figure}[!htb]
\centering
\includegraphics[width=0.485\textwidth]{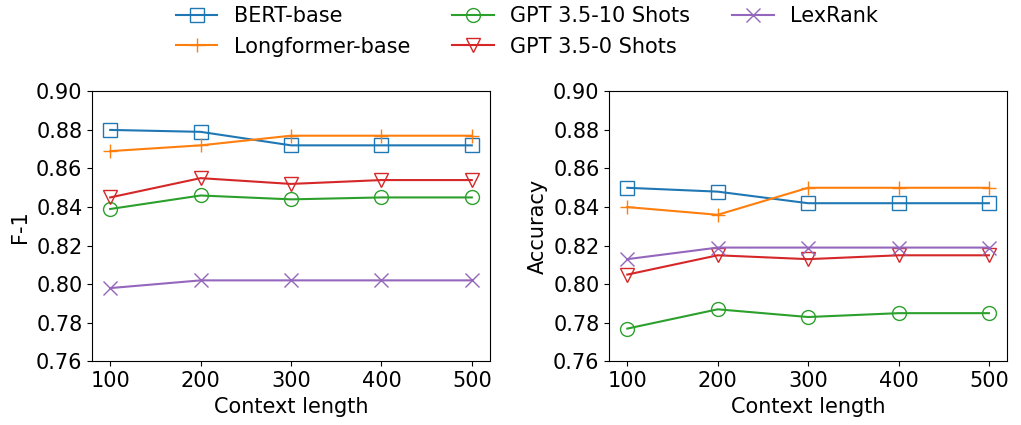}
\caption{Model performance when using a context collected from biased news sources and summarized in 500 words then trimmed at different lengths.}
\label{fig:biased_trimmed_context}
\end{figure}

{\flushleft{\textbf{Biased non-neutral context}}}\hspace{2mm} 
In addition to context collected from least-biased sources, we experimented with context from biased sources.
Particularly, for each event we collected context from a left-biased source and a right-biased source.\todo[color=green]{It is unclear what "different biases selected randomly" means. Are the "biases" randomly selected"? Or the sources are randomly selected, for each bias? The biases are only qualified (instead of quantified" as left or right, correct? If you randomly select, does it mean the two sources can be both left?} We then concatenated and fed the two articles to a generative LLM, specifically GPT 3.5, to summarize them in a single text of varying lengths. The reason for summarizing the two articles is to make sure their gist fits the upper limits on the input sequence length of the different models.\todo[color=green]{Unclear whether we summarize the two articles together or separately.} We ran all models against the data sets with summarized contexts. Additionally, we asked the generative LLM to summarize the context collected from biased sources in 500 words, however, this time we trimmed the 500 words summary at different lengths.\todo[color=green]{Which articles are you referring to? the two articles from biased sources? is this last sentence repeating what's said before it?} We then ran all models against the data sets with summarized-then-trimmed contexts.\todo[color=green]{Again, is this repeating what's said in `` We ran all models against the data sets with summarized contexts.''} In these two experiments (i.e., summarized in different lengths vs. summarized at 500 words then trimmed)\todo[color=green]{Which two experiments?} we used the same settings and hyper-parameters as in the previous experiments. 
Results in Figure~\ref{fig:biased_context} show that our proposed methodology does not rely on the existence of a neutral context to perform well. Instead, context articles from two sources with different biases can neutralize each other and lead to performance comparable to the least-biased context. Note how performance becomes more stable when context is summarized then introduced to the models at different lengths as Figure~\ref{fig:biased_trimmed_context} shows.\todo[color=green]{Unclear to me what ``lose effect'' means. Actually, it appears the best performance was obtained with shortest context.} One explanation is that the most important statements are present in the beginning of an article and thus its summary. This means that once a model sees all the important statements within the context its performance either stops improving or degrades as it gets distracted by additional insignificant text.

\vspace{-1mm}

{\flushleft{\textbf{Importance granularity.}}}\hspace{2mm} We anticipate that the performance of models trained on varying levels of statement importance granularity will differ. For instance, we expect supervised models to be more precise in drawing a boundary between the labels \textit{very important} versus everything else, compared to drawing a boundary between more fine-grained importance labels such as \textit{very important} versus \textit{kind of important}. To examine our models' capabilities across different levels of importance granularity, we trained our top-performing model (i.e., Longformer-large) using the four classification configurations outlined in Table~\ref{table:class_distrib}. The results in Table~\ref{table:longformer-results} show that the difficulty of learning a decision boundary decreases when classes are merged and thus label granularity decreases. This trend is particularly notable in the case of \textit{kind of important} and \textit{not very important}. When these two classes are separate (configuration \#4 in Table~\ref{table:class_distrib}), the performance is significantly inferior, compared to other configurations where these two classes are merged.
This is attributed to the high fuzziness between the two classes. Therefore, we utilize configuration \#1 from  Table~\ref{table:class_distrib} in all remaining experiments.

\begin{table}[!htb]
\small
\centering
\begin{tabular}{|c|c|c|} 
 \hline
\textbf{Conf. \# } & \textbf{Acc.} & \textbf{F-1} \\ 
 \hline
 1 &0.897 & 0.887\\ 
 \hline
 2 & 0.853 & 0.828 \\
 \hline
 3 & 0.875 & 0.898 \\
 \hline
 4 &  0.728 & 0.717\\
 \hline
\end{tabular}
\caption{Performance of the Longformer model in the four classification configurations.}
\label{table:longformer-results}
\end{table}
\vspace{-1mm}

{\flushleft{\textbf{End-to-end cherry-picking detection.}}}\hspace{2mm} To assess the performance of the complete cherry-picking detection pipeline, we examined the relationship between two variables, $x$: Cherry-picking score calculated on the detected cherry-picking by our automatic pipeline in each news outlet and $y$: the bias scores of the outlet collected from MBFC and AllSides.\todo{Earlier we also mentioned Ad Fontes Media. Why is it left out? Do we need to explain?} Our hypothesis is that, since cherry-picking is a form of bias, there should be some level of correlation between the two variables. 

To test the relationship between variables $x$ and $y$, we ran the full cherry-picking detection pipeline described in Formula \ref{eq1} employing our top-performing supervised model on 2,453 events comprised of about 97k statements collectively. We ensured that these events did not overlap with the Cherry dataset utilized for training our supervised models. Next, we calculated the average number of cherry-picked statements per outlet across all events, resulting in a final score of $x$. Finally, since variable $y$ is ordinal, we calculated the Spearman's correlation coefficient $r$ between $x$ and $y$. Results in Table~\ref{table:correlation} show a positive moderate correlation approaching statistical significance when the bias scores come from AllSides.com. Additionally, there is a positive weak quasi-significant correlation when the bias scores come from MBFC.~\footnote{Further insights into the interpretation of correlation coefficient, including what is considered moderate and what is weak, can be found in~\cite{ratner2009correlation}.}

A strong correlation between the two variables is not evident. It could be due to a few reasons. First, the bias scores provided by MBFC and AllSides were calculated by assessing multiple bias forms including those beyond cherry-picking, e.g., lexical cues embedded in text. Furthermore, MBFC and AllSides continually update their bias scores using fresh article samples. On the country, the inference of our models was applied on static data from each outlet collected between December 2019 and January 2021. 

\begin{table}[!htb]
\small
\centering
\begin{tabular}{|c|c|c|} 
 \hline
\textbf{Bias score source} & \textbf{r} & \textbf{P-value} \\ 
 \hline
 MBFC & 0.28 & 0.10\\ 
 \hline
 AllSides & 0.32 & 0.06 \\
 \hline
\end{tabular}
\caption{Correlation between $x$: cherry-picking detection pipeline scores and $y$: bias scores from MBFC and AllSides.}
\label{table:correlation}
\end{table}
\vspace{-1mm}

Lastly, we hypothesize that the average cherry-picking score $x$ for biased outlets, belonging to a particular bias band, should align with their bias intensity. For instance, we anticipate observing a higher average cherry-picking score among outlets categorized as ``left bias'' compIared to those classified as ``left-center bias.'' To test this, we calculated the mean and standard deviation of the cherry-picking score of all outlets under each bias category as Table \ref{table:bias_bands} shows. Results show an uptrend in cherry-picking scores that aligns with the intensity of bias in the analyzed outlets. Nevertheless, the observed pattern is affected by the limited sample sizes, highlighting the need for future analysis involving larger samples.

\begin{table}[!htb]
\small
\centering
\begin{tabular}{|c|c|c|c|} 
 \hline
\textbf{Bias category} & \textbf{Mean} & \textbf{STD} & \textbf{Sample size} \\ 
 \hline
 Left & 15.12 & 4.39 & 6\\ 
 \hline
 Left center & 10.32 & 3.05 & 15 \\
 \hline
 Right & 8.91 & 4.09 & 7 \\
 \hline
 Right center & 8.33 & 1.21 & 5 \\
 \hline
 Center & 8.44 & 0.30 & 2 \\
 \hline
\end{tabular}
\caption{The mean and standard deviation of cherry-picking scores aggregated by bias category.}
\label{table:bias_bands}
\end{table}
\vspace{-1mm}

\section {Conclusions and Future Work}
\label{sec:conclusion}
Manually spotting cherry-picking statements in news reports is challenging due to the need to examine other narratives. This study introduces a novel approach to automate the detection of cherry-picking in news reports. Our approach focuses on comparing multiple news articles that pertain to the same event and identifying the omission of crucial statements. To facilitate our research, we have constructed the first cherry-picking detection dataset. The results of our models demonstrate promising outcomes. Currently, our work specifically addresses the detection of cherry-picking based on underemphasis of statements. For future work, we plan to also considering overemphasis. 



\section*{Limitations}
\label{sec:limitation}
\stitle{Dependence on semantic similarity thresholds and adjustable parameters.} 
A notable limitation of the end-to-end cherry-picking pipeline, as outlined in Section \ref{sec:model}, is its dependence on semantic similarity thresholds. This can be observed in the application of clustering techniques during the creation of events from a set of articles. If a news source covers an event through multiple articles and one of these articles is not clustered into the correct event, the pipeline may overlook important information from the coverage of that outlet. Moreover, when determining whether an article mentions a specific statement, if the threshold for semantic similarity is not well-tuned, similar statements in the article may not be detected and consequently be falsely considered cherry-picked.\\
\stitle{Definition and tokenization of statements.} 
To ensure the precise identification of cherry-picked statements, it is essential to have a clear definition and an appropriate tokenization method for segmenting text into statements. Currently we consider every sentence as a statement, but this approach may not always be accurate since multiple sentences can collectively form a single statement.

\section*{Ethics Statement}
While our work focuses on detecting bias by omission, it inevitably incorporates a human element in two stages. First, At the data collection stage, human coders are asked to indicate whether a statement is important to the event. This may result in a dataset that contains the inherent subjectivity and personal perspectives of the human annotators themselves. We are aware that annotators' implicit biases can influence the annotations they provide. Therefore, we tried to alleviate this influence by hiding sources names from segments in the annotation tasks, in addition to providing annotators with context to read before they assign labels. Second, bias categorizations we use from MBFC, AllSides, and Ad Fontes Media rely on human annotators as well. Hence, the bias of these sources can be inherently embedded into our dataset. While we hold the belief that the sources we rely on exhibit high quality, it is important to recognize that their reliability and credibility may decline due to various unforeseen factors.
At last, we assert that our usage of all artifacts this work including pretrained models (e.g., Longformer and BERT) and derived data (e.g., GDELT events data and labels from MBFC, Ad Fontes Media and AllSides) is purely for research purposes and consistent with their intended use.

\bibliography{anthology,custom}
\bibliographystyle{acl_natbib}

\appendix
\renewcommand{\thesection}{Appendix \Alph{section}}

\section{Cherry-picking Example}
\label{sec:juxta_example}
Table \ref{table:juxtaposition} illustrates cherry-picking in journalism by showcasing statements from authentic news articles by distinctively biased news outlets, all covering the same event: U.S. President Donald Trump's attendance at the annual Davos economic forum. The three news outlets included in Table~\ref{table:juxtaposition} are Reuters.com (center), NewsMax.com (right-bias), and CNBC.com (left-bias). The bias categorization of these media outlets is determined by assessments from Media Bias Fact Check (MBFC). The table  organizes corresponding statements from the three articles in rows based on their semantic similarity. If a statement is not mentioned by a particular news source, the corresponding cell in the table is left empty. Note that Reuters.com and NewsMax.com published identical articles for this event. \todo[color=pink]{Ideally you need a different example with three different articles.}

By examining Table \ref{table:juxtaposition}, we can observe instances of cherry-picking. For example, in the third row, two of the news outlets mentioned that President Trump attended the Davos forum last year (i.e., 2019). However, CNBC.com (left-bias) failed to mention it, and focused instead on details about other issues including placing tariffs on cars imported from European Union and French specialties, as shown in the eighth row of the table. 

\begin{table*}[h]
\small
\centering

\begin{tabular}[hbt!]{p{0.47\linewidth}| p{0.47\linewidth}}
\hline

      \textbf{Reuters.com (Center) }/ \textbf{NewsMax.com (Right-bias)} &   \textbf{CNBC.com (Left-bias)} \\ \hline \hline
U.S. President Donald Trump plans to attend the annual Davos economic forum in January. &
After skipping a gathering of the world's most elite in Davos, Switzerland last January, President Donald Trump will attend the World Economic Forum in 2020.\\ \hline
Trump had to cancel his plan to attend the annual gathering of global economic leaders early this year due to a government shutdown.&
Trump blamed his no-show last time on the partial government shutdown that was triggered by a funding dispute over a proposed wall along the United States' southern border. \\ \hline
He attended the Davos forum last year.&
- \\ \hline
The exact date of when Trump would participate was unclear. &
A spokesperson for the White House did not immediately respond to a request for comment about the president's plans to attend next year.\\ \hline
Davos may be one of the few foreign trips that Trump takes in 2020.&
- \\ \hline
The World Economic Forum in the Swiss ski resort town of Davos is scheduled to run January 21-24. &
The 50th annual forum, held Jan. 21 - Jan. 24 will focus on ``Stakeholders for a Cohesive and Sustainable World.''\\ \hline
-&
Meantime, while the Davos agenda will look to ``create bridges to resolve conflicts in global hotspots,'' Trump has made jabs at a number of European allies.  \\ \hline
-&
While he has said he was just joking about placing tariffs on cars imported from the European Union, he is still threatening to place tariffs on French specialties like champagne and cheese. \\

\hline 
\end{tabular}
\caption{A juxtaposition of three news stories covering the same event, published by three sources with different political biases. Bias categorization of these media outlets is based on MediaBiasFactCheck.com (MBFC).} 
\label{table:juxtaposition}
\end{table*}
\label{sec:appendix}

\section{Media Bias Rating Sources}
\label{sec:rating_method}
MBFC, AllSides, and Ad Fontes Media employ diverse methods to evaluate bias in news outlets.
Literature addressing bias and misinformation detection heavily relies on these three sources for bias categorization in tasks such as social media source annotation~\cite{weld2021political}, comparing the diffusion of news from reliable and non-reliable sources~\cite{cinelli2020covid}, and benchmarking automatic media bias monitors~\cite{ribeiro2018media}.


\stitle{MBFC} utilizes a comprehensive methodology to assess the ideological leanings and factual accuracy of media outlets.\footnote{\url{https://mediabiasfactcheck.com/methodology/}} The bias assessment includes categories such as Biased Wording/Headlines, Factual/Sourcing, Story Choices, and Political Affiliation. The scoring mechanism categorizes sources into Least Biased, Left/Right Center Bias, Left/Right Bias, or Extreme Bias. The methodology also assesses factual reporting with ratings ranging from Very High to Very Low based on the reliability and commitment to factual accuracy. 

\stitle{AllSides} employs a comprehensive methodology to rate media bias, including various factors and perspectives.\footnote{\url{https://www.allsides.com/media-bias/media-bias-rating-methods}} AllSides Editorial Reviews involve a multipartisan panel assessing news reports and bias indicators. Blind Bias Surveys gather opinions from average Americans across political spectrums. AllSides reviewers assess content independently, considering common bias indicators and transparency about political leaning. AllSides also incorporates third-party data and allows Community Feedback for additional perspectives. 

\stitle{Ad Fontes Media} generates scores for news sources based on the ratings of individual articles or episodes, using over 60 trained analysts from diverse backgrounds who undergo initial and ongoing training.~\footnote{\url{https://adfontesmedia.com/how-ad-fontes-ranks-news-sources/}} Each source is rated by at least three analysts with different political leanings. The analysts independently rate the content, compare scores, and discuss any discrepancies. The overall rating is the average of the analysts’ ratings. In some cases, more analyses may be used for rating articles with outlier scores.

\section{Clustering of $S_e$}
\label{sec:clust_example}


Table~\ref{table:clustering} shows two statement clusters from two different events.

\begin{table*}[hbt]
\small
\centering
\begin{tabular}{|p{6.0in}|} 
 \hline
\textbf{Cluster \#1} \\ 
 \hline
 $\cdot$ President-elect Joe Biden plans to release nearly all available doses of the COVID-19 vaccines after he takes office.\\ 
 $\cdot$ President-elect Joe Biden plans to release almost all vaccine doses immediately.\\
 $\cdot$ President-elect Joe Biden will aim to release every available dose of the coronavirus vaccine when he takes office.\\
 $\cdot$ Joe Biden will release most available Covid-19 vaccine doses to speed delivery to more people when he takes office. \\
 \hline
\textbf{Cluster \#2} \\ 
 \hline
 $\cdot$ The House voted to override President Trump’s veto of a \$740 billion defense spending and policy bill.\\
 $\cdot$ The House of Representatives on Monday voted to override President Trump’s veto of the National Defense Authorization Act for Fiscal Year 2021.\\
 $\cdot$ House Votes to Override Trump’s Veto of 2021 Defense Policy Bill.\\
 $\cdot$ The House voted late Monday to override President Donald Trump’s veto of a defense spending bill for 2021.\\
 \hline
\end{tabular}
\caption{Two examples of statement clusters.}
\label{table:clustering}
\end{table*}

\begin{figure*}[hbt]
\centering
\includegraphics[width=1\textwidth]{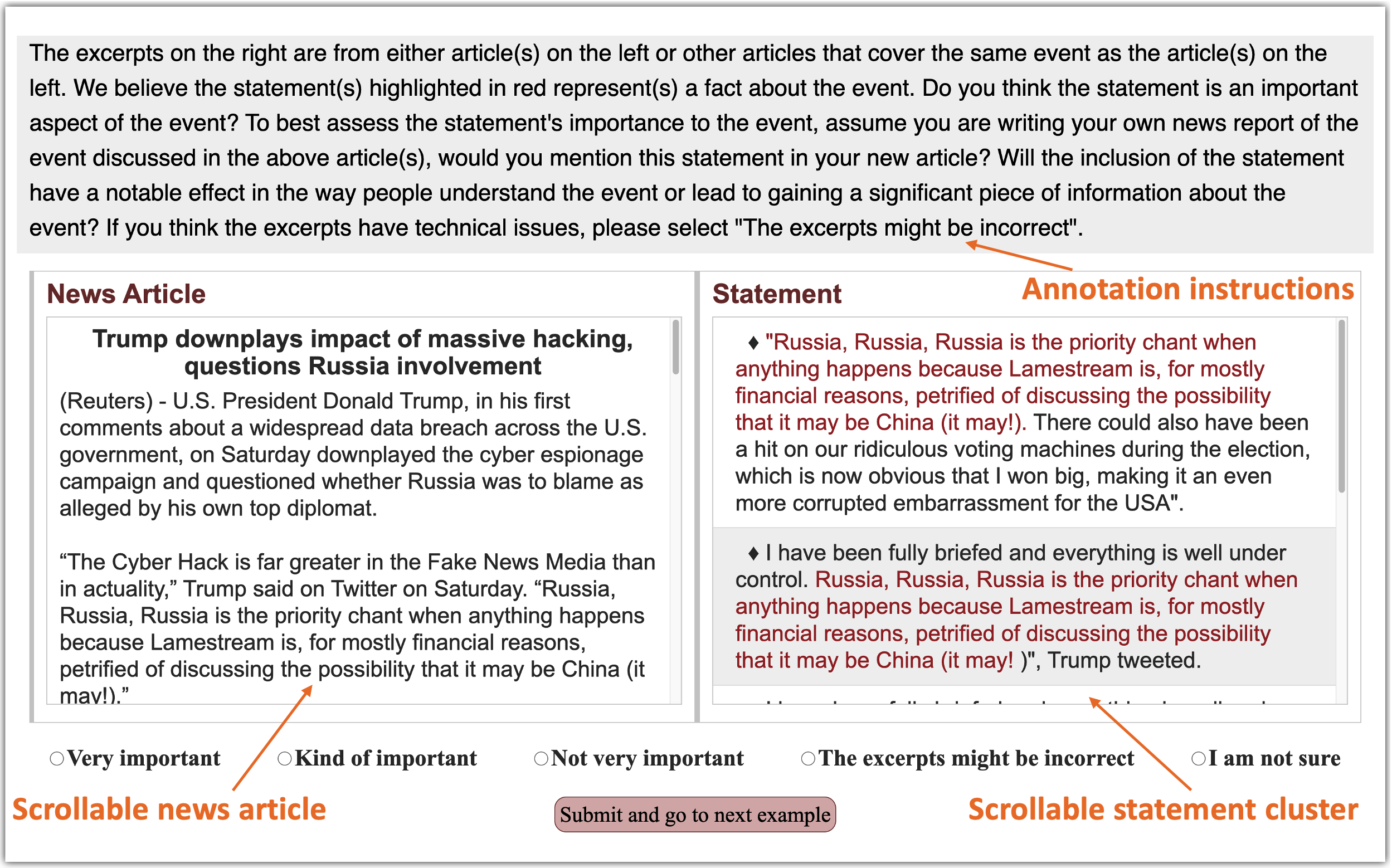}
\caption{Cherry-picking data annotation interface.}
\label{cherry_web}
\end{figure*}

\section{Annotation Interface}
\label{sec:interface} 
Upon logging into the interface, the annotators enter an assigned event ID. The annotators are instructed to assign one of five labels to the statement cluster based on their understanding of the event described in the news article (Figure~\ref{cherry_web}). 
A statement is considered important if it articulates a fact essential to understanding the event, and vice versa. If the statement cluster is unrelated to the event, lacks meaningfulness, or contains facts with conflicting meanings, the annotators are expected to label it as (4) \textit{the excerpts might be incorrect}.  The annotators can select (5) \textit{I am not sure} if they feel uncertain about the statements' importance. After selecting a label, the annotators can proceed by clicking the ``Submit and go to next example'' button to annotate the subsequent statement clusters associated with the same event. Once the annotators completes all statement clusters of the event, they will receive a prompt to enter a new event ID and proceed to the next event.

\section{Importance Assessment as a Sequence-pair Classification Task}
\label{sec:archi}

\begin{figure}[!htb]
\centering
\includegraphics[width=0.35\textwidth]{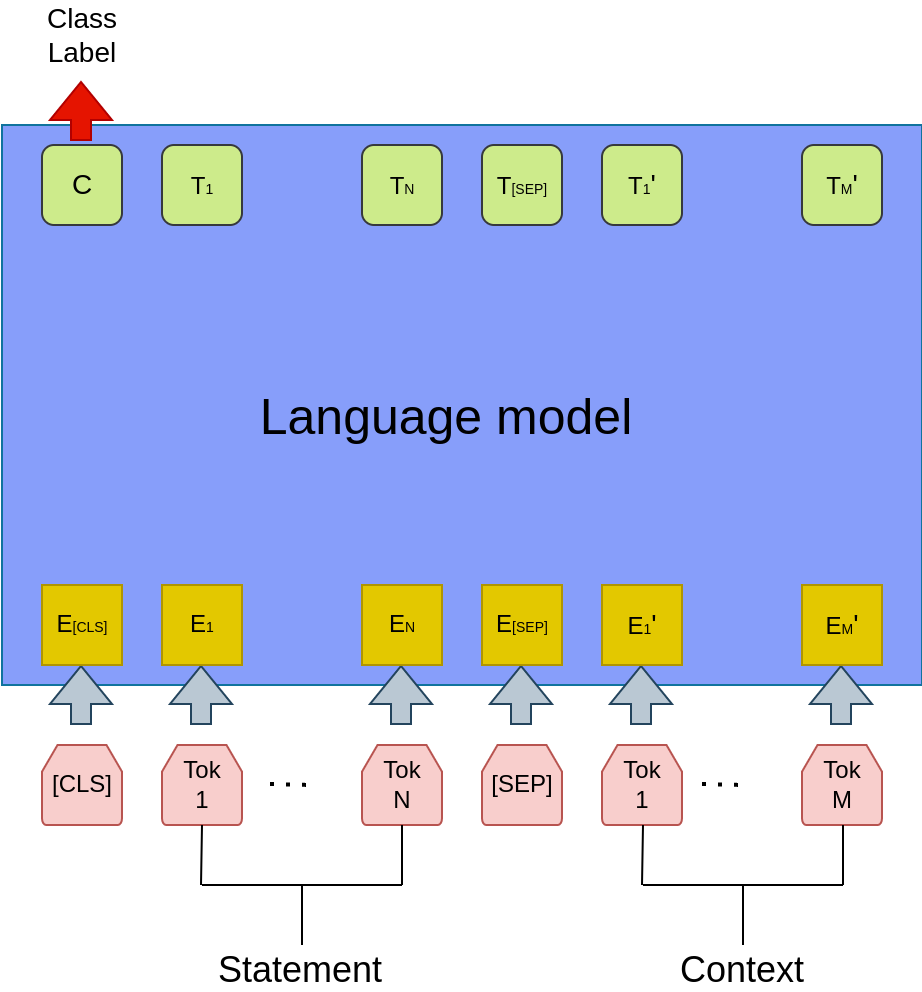}
\caption{Automatic cherry-picking detection using sequence-pair classification architecture.}
\label{fig:archi}
\end{figure}

The supervised BERT and Longformer models are trained using the sequence-pair classification task, with each input consisting of a statement and the corresponding context derived from the given event, as shown in Figure~\ref{fig:archi}. The output embedding vector of the classification token is softmax--ed to produce the positive class probability.

\begin{figure*}[hbt]
\centering
\includegraphics[width=0.90\textwidth]{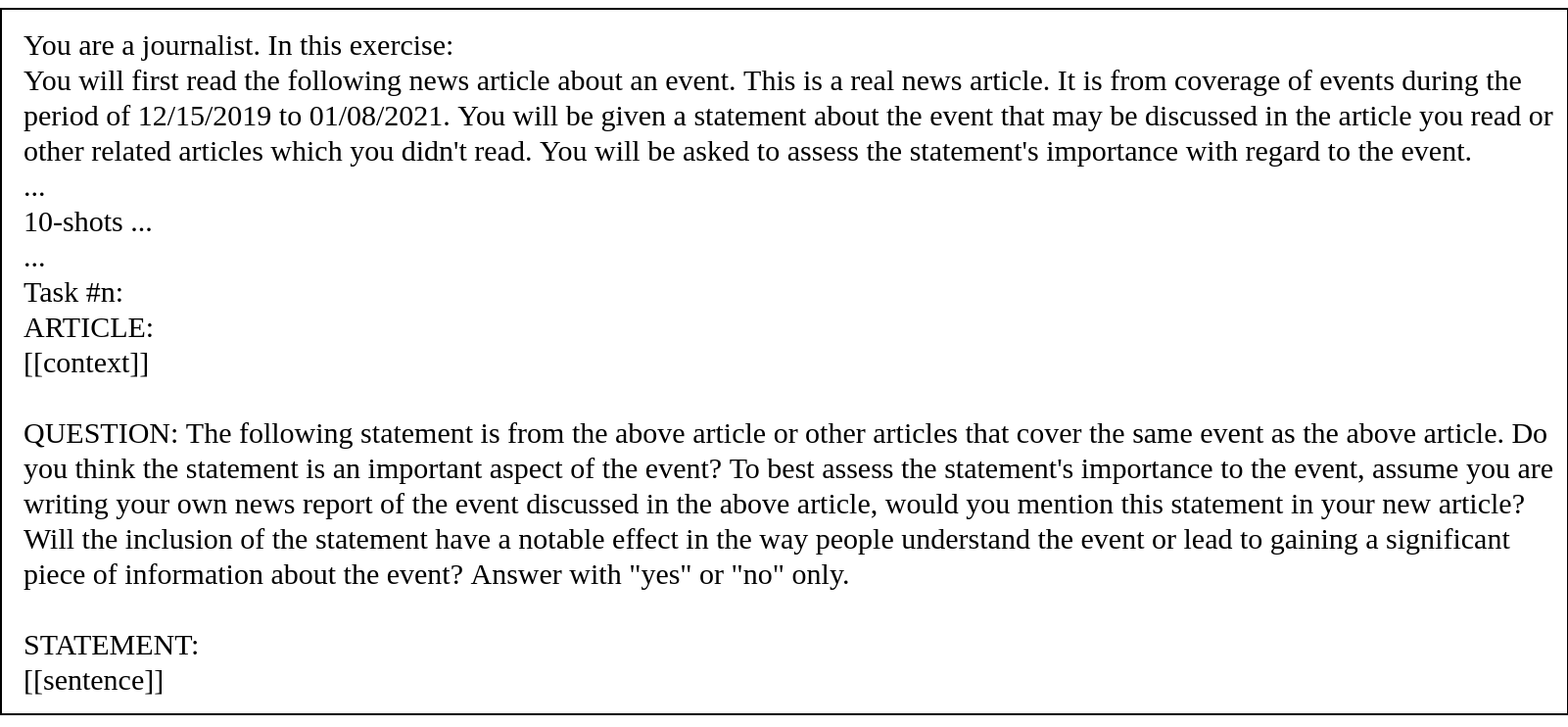}
\caption{Template used to prompt GPT.}
\label{fig:prompting_template}
\end{figure*}

\section{Global Attention in Longformer}
\label{sec:glob_attention}
To evaluate the influence of the global attention mechanism on predicting a statement's importance, we tested our best model (Longformer-large) with all possible combinations of global attention locations within the input sequence, i.e., the classification token, the statement tokens, and the context tokens. We maintained the sequence length of 512 and set the batch size to 4 in this experiment to ensure memory can handle applying global attention on all the sequence tokens. Results in Table \ref{table:global_attention} show that we can still maintain the same performance but optimize memory and inference time~\cite{beltagy2020longformer} by applying global attention on the statement tokens and classification tokens only. For this reason, we fixed global attention to these two locations in all of our experiments. Moreover, we can optimize further and trade off more efficiency at the expense of a slight decrease in performance by applying global attention on the classification token only. \todo[color=pink]{having experiment results about memory consumption and training/inference time will make this setion more convincing.}

\begin{table}[!htb]
\small
\centering
\begin{tabular}{|c|c|c|} 
 \hline
\textbf{Global attention locations } &  \textbf{Acc.} & \textbf{F-1} \\ 
 \hline
 [CLS] & 0.819 & 0.820\\ 
 \hline
 [CLS] + statement & 0.831 & 0.837 \\
 \hline
 [CLS] + context & 0.683 & 0.680\\
 \hline
 [CLS] + statement + context & 0.837 & 0.838 \\
 \hline
\end{tabular}
\caption{Performance of the Longformer-large model using different global attention locations.}
\label{table:global_attention}
\end{table}

\section{GPT Prompts}
\label{sec:prompting}
To prompt GPT, we experimented with various prompts, including simple ones such as the question \textit{``Is the above sentence important to mention in a news article that covers the story mentioned in the above article? Answer with `yes' or `no' only.''} Using the prompt as shown in Figure~\ref{fig:prompting_template}, which is crafted based on the annotation interface, yielded the best GPT-based performance on the test dataset. 


\todo[color=green]{For the prompt figure, would it be better to make this landscape instead of portrait? Also, the structure will be come more clear if certain words are highlighted. isRAA: fixed}


\end{document}